%% file: main.tex
\documentclass[10pt,twocolumn,letterpaper]{article}

\usepackage[pagenumbers]{cvpr}

\input{preamble}
\definecolor{cvprblue}{rgb}{0.21,0.49,0.74}
\usepackage[pagebackref,breaklinks,colorlinks,allcolors=cvprblue]{hyperref}
\usepackage{tabu}                      
\usepackage{booktabs}                  
\usepackage{lipsum}                    
\usepackage{mwe}                       
\usepackage{subcaption}     
\usepackage{enumitem}
\usepackage{wrapfig}
\usepackage{url}            
\usepackage{amsfonts}       
\usepackage{nicefrac}       
\usepackage{amsmath}
\usepackage{microtype}      
\usepackage{xcolor}         
\usepackage{acronym}
\usepackage{multirow}
\usepackage{xspace}
\usepackage{graphicx}       
\usepackage{enumitem}
\usepackage{wrapfig}
\usepackage{placeins}
\usepackage{caption}
\usepackage{cuted}

\renewcommand{\footnotemark}{\relax}
\title{Seeing My Future: \\ Predicting Situated Interaction Behavior in Virtual Reality}

\author{
    Yuan Xu\textsuperscript{1,*} \quad
    Zimu Zhang\textsuperscript{1,*} \quad
    Xiaoxuan Ma\textsuperscript{1} \quad
    Wentao Zhu\textsuperscript{2} \quad
    Yu Qiao\textsuperscript{3} \quad
    Yizhou Wang\textsuperscript{1}
    \\[1.5ex]
    \textsuperscript{1} Peking University \quad
    \textsuperscript{2} Eastern Institute of Technology, Ningbo \quad
    \textsuperscript{3} Shanghai Jiao Tong University \\
}

\begin{document}

\maketitle
\footnote{* Equal contribution.}

\input{sec/notation}
\input{sec/0_abstract}
\input{sec/1_introduction.tex}
\input{sec/2_related_work.tex}
\input{sec/3_method}

\input{sec/4_experiment.tex}

\input{sec/5_conclusion}

{
    \small
    \bibliographystyle{ieeenat_fullname}
    \bibliography{main}
}
\input{sec/X_suppl}

\end{document}

%% file: sec/0_abstract.tex
\begin{figure*}[h!]
  \centering
  \includegraphics[width=\linewidth]{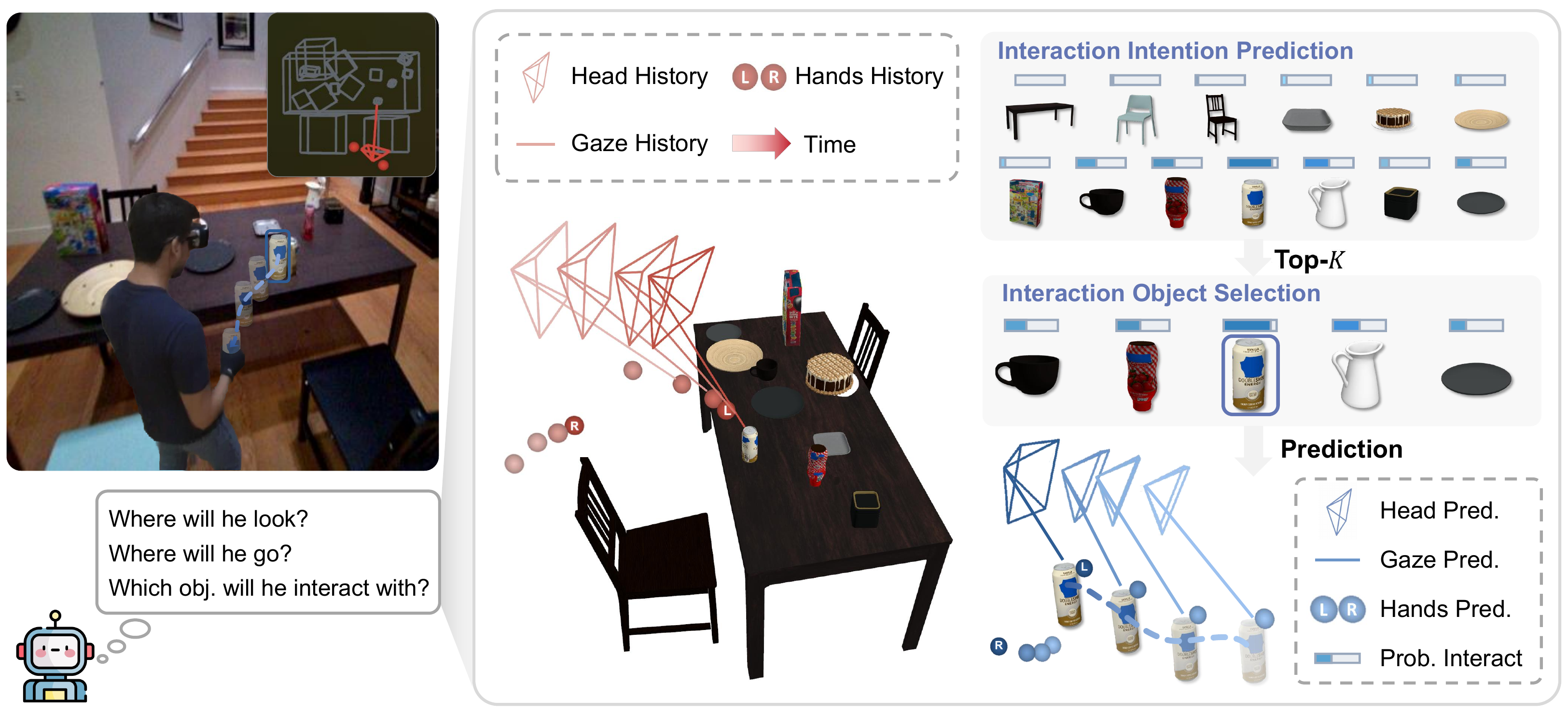}
  \caption{\textbf{Situated interaction behavior prediction}.
  Given historical human dynamics, such as gaze direction and situational context, we propose an intention-aware framework to predict a person’s future behavior in the environment, including where to look (\emph{gaze}), where to go (\emph{trajectory}), and which objects to interact with (\emph{object interaction}). Our approach employs a hierarchical prediction strategy aligned with human cognition, first coarsely identifying potential objects based on prior engagements before precisely forecasting the next specific interaction.}
  \label{fig:teaser}
  \vspace{-0.4cm}
\end{figure*}

\begin{abstract}
Virtual and augmented reality systems increasingly demand intelligent adaptation to user behaviors for enhanced interaction experiences. Achieving this requires accurately understanding human intentions and predicting future situated behaviors—such as gaze direction and object interactions—which is vital for creating responsive VR/AR environments and applications like personalized assistants. However, accurate behavioral prediction demands modeling the underlying cognitive processes that drive human-environment interactions. 
In this work, we introduce a hierarchical, intention-aware framework that models human intentions and predicts detailed situated behaviors by leveraging cognitive mechanisms. Given historical human dynamics and the observation of scene contexts, our framework first identifies potential interaction targets and forecasts fine-grained future behaviors. We propose a dynamic Graph Convolutional Network (GCN) to effectively capture human-environment relationships.
Extensive experiments on challenging real-world benchmarks and live VR environment demonstrate the effectiveness of our approach, achieving superior performance across all metrics and enabling practical applications for proactive VR systems that anticipate user behaviors and adapt virtual environments accordingly.
\end{abstract}

%% file: sec/1_introduction.tex
\section{Introduction}
\label{sec:intro}

Anticipating and responding to user interactions represents the next frontier for creating more immersive and effective AR/VR experiences. Moving beyond reactive approaches, proactive adaptive systems promise to revolutionize applications spanning personalized gaming, training simulations, collaborative workspaces, and assistive technologies. To realize these systems, accurately predicting situated behaviors—where action arises from the interplay of intention and context—remains essential for enabling intelligent, adaptive systems across diverse applications.

For example, in virtual environments, predicting situated behavior enables VR systems or immersive games to adapt environment complexity or interaction modes in response to user intent, creating more engaging and personalized experiences~\cite{nasri2025towards, wang2024predicting, zhang2025facilitating}. A VR game might, for instance, predict which area the player is about to explore and introduce ambient story elements or adjust enemies accordingly.
In the physical world, such predictive capabilities allow robots and AR systems to interpret user intent in context, offering proactive support for smoother human–AI collaboration~\cite{belcamino2024gaze, li2025satori, mascaro2023hoiabot}.

Achieving such accurate behavioral prediction requires a deeper understanding of the cognitive processes underlying human–environment interactions. Human interaction behaviors are not random but emerge from underlying intentions that are continuously shaped and reshaped by dynamic situational contexts, such as environmental cues, task demands, and temporal constraints~\cite{ALBARRACIN2024151, tong2022context, tong2025human, triberti2016being}. Research in psychology and cognitive science \cite{brand1984intending, bratman1987intention, mele1992springs, pacherie2008phenomenology}~ suggests that humans often first form an interaction intention (\eg, identifying potential objects for interaction), which then guides the planning and execution of specific actions. In the formation of interaction intention, the relationship between gaze and the environment plays a crucial role \cite{land2006eye}. Gaze, by enabling the active search for relevant information in the environment, serves as the most direct and observable expression of attention \cite{belardinelli2024gaze}. It often precedes and guides interactions by locking onto targets that align with an individual’s goals \cite{johansson2001eye, pelz2001oculomotor, land2006eye}.

In light of this, we propose an intention-aware framework to predict comprehensive human situated behavior in an environment, including where they will look (\emph{gaze}), where they will go (\emph{trajectory}), and which objects they will interact with (\emph{object interaction}), as shown in \cref{fig:teaser}.
The core of the proposed framework lies in a hierarchical modeling approach, inspired by the human cognitive process. Initially, it predicts a coarse set of potential interaction targets, such as a ketchup bottle or cake on a table based on historical gaze attention and human-scene interactions (see \cref{fig:teaser}). From these, it identifies the top-$\topk$ objects with the highest interaction likelihood. Subsequently, the framework forecasts detailed future behaviors, specifying both the object of interaction (\eg, a coffee can as the final target) and the corresponding action, such as picking up the coffee can, as illustrated in \cref{fig:teaser}. This predictive process is powered by a dynamic Graph Convolutional Network (GCN) to effectively capture the human-environment interactions and enhance the accuracy and robustness of behavior prediction in dynamic immersive environments.

To validate the practical applicability of our framework, we conduct user studies in a VR environment where participants engage in natural interaction sequences. Our results demonstrate robust performance even under noisy, realistic conditions, showing that the system can successfully anticipate user gaze patterns and future interaction behaviors, and highlighting the framework's potential to support a new generation of proactive VR applications.

To summarize, our contributions are three-fold: 
\begin{itemize}[leftmargin=*, align=left]
 \item  We propose a hierarchical and intention-aware framework to model the underlying human cognitive process to help accurately predict situated behavior in the environment.
 \item  We design a dynamic GCN with an adaptive weight matrix that effectively captures context-aware relationships between humans and the environment. 
 \item  We conduct an empirical validation of the framework's practical applicability through an experiment in a real-world VR environment, demonstrating its robustness and effectiveness under realistic, noisy conditions.
\end{itemize}

\begin{figure*}[t] 
  \centering
  \includegraphics[width=\textwidth]{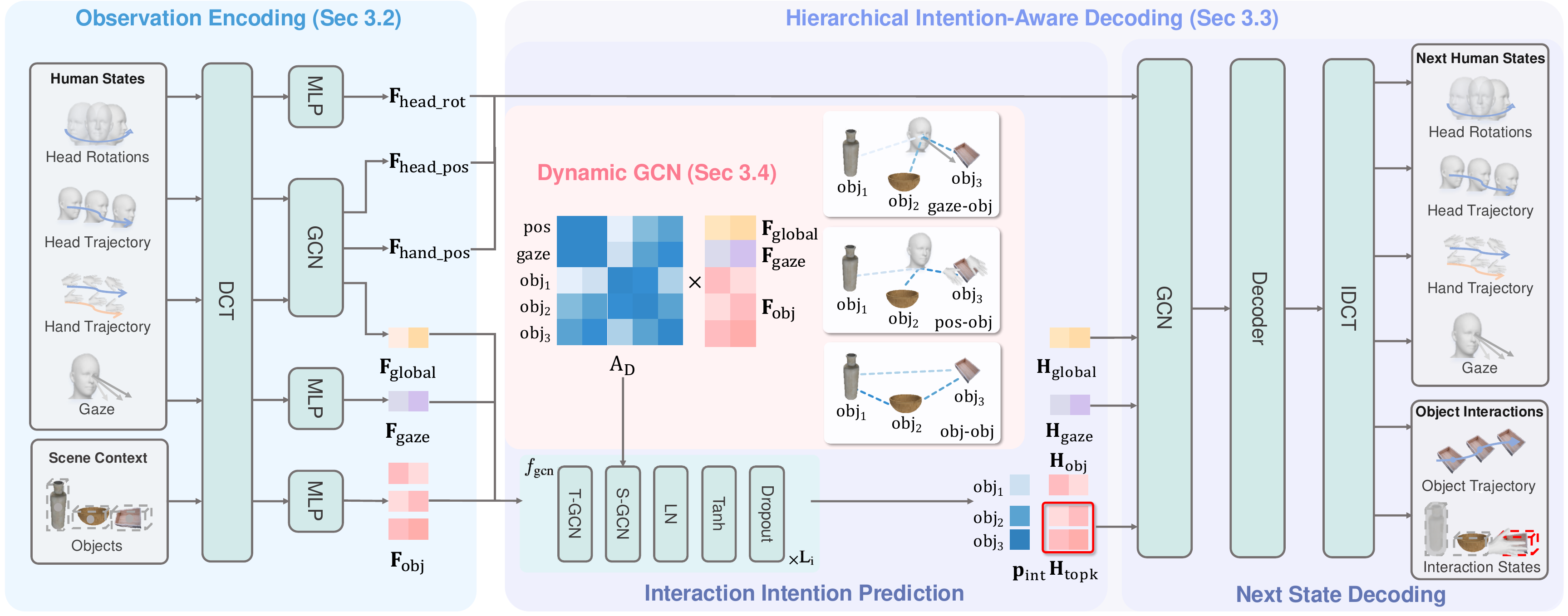}
  \caption{\textbf{Overview of our framework.} (1) an observation encoding module that captures and encodes historical human states and scene context, (2) a hierarchical intention-aware decoding module that first predicts potential top $\topk$ interaction targets and then forecasts detailed human next states as well as the object interactions, and (3) a dynamic GCN that adaptively models relationships among human gaze, head positions, hand positions, and objects.} 
  \label{fig:pipeline} 
  \vspace{-0.4cm}
\end{figure*}

%% file: sec/2_related_work.tex
\section{Related Work}

\noindent\textbf{Gaze-body Correlation.} The correlation between gaze and body movements during human-object interactions has been extensively explored in prior research \cite{johansson2001eye, pelz2001oculomotor, batmaz2020touch, coudiere2024eye, bertrand2023dynamics, de2022eye, de2018keeping, fang2015eye, hu2019sgaze, pfeil2018comparison, stamenkovic2018postural, kothari2020gaze, hu2024pose2gaze, sidenmark2019eye, zheng2022gimo, lou2024multimodal}, which can be broadly categorized into three main strands. The first line of works focuses on the correlation between gaze, hands, and objects \cite{johansson2001eye, pelz2001oculomotor, batmaz2020touch, coudiere2024eye, bertrand2023dynamics, de2022eye, de2018keeping}. They find that gaze typically pre-plans the trajectory for hand movements, while gaze fixations selectively focus on objects that are highly relevant to the current task. The second line of works analyzes gaze-head coordination \cite{fang2015eye, hu2019sgaze, sidenmark2019eye, pfeil2018comparison, stamenkovic2018postural, kothari2020gaze}. One of their main findings is the correlation among gaze position, head orientation, and head rotation velocity. Moreover, they also reveal that gaze movements slightly precede head movements. The third line of works explores the correlation between gaze and the full-body \cite{hu2024pose2gaze, sidenmark2019eye, zheng2022gimo, lou2024multimodal}. For example, Sidenmark \etal \cite{sidenmark2019eye} examine the patterns between gaze shifts and body movements and their temporal alignment. Hu \etal \cite{hu2024pose2gaze} analyze the correlation between gaze movements and human body movements. Inspired by the discovery that gaze plays a pivotal role in human-object interactions, offering valuable cues for predicting human actions, we leverage gaze information to enhance the accuracy of human-object prediction as well. \\

\noindent\textbf{Gaze Prediction.} Gaze, as an explicit representation of human visual attention, offers important information for modeling human intentions and behavior \cite{hu24gazemotion, zheng2022gimo, lou2024multimodal, ozdel2024gaze, yan2023gazemodiff, grand2025gaipat}. Existing works on gaze prediction can be categorized into two groups: 2D gaze prediction \cite{fathi2012learning, huang2018predicting, hu2019sgaze, lai2024eye, kim2022memory, lai2024listen} and 3D gaze prediction \cite{hu24gazemotion, hu2024pose2gaze, hu2022we, kellnhofer2019gaze360, wei2018and}. 2D gaze prediction aims to estimate gaze positions in egocentric videos. Pioneer work \cite{fathi2012learning} proposes the problem of predicting gaze from a 2D egocentric video. Huang \etal \cite{huang2018predicting} tackle this problem by modeling the attention transition and predicting the saliency map for each frame. Recent work \cite{lai2024eye} leverages a transformer-based model to enhance performance. For 3D gaze prediction, the goal is to predict the gaze in a 3D scene. For example, GazeMotion \cite{hu24gazemotion} tries to use a CNN-based model to predict future gaze vectors and then uses the predicted gaze to help human motion prediction. Pose2Gaze \cite{hu2024pose2gaze} utilizes the historical human pose feature to predict 3D gaze. Wei et al. \cite{wei2018and} further explored the cognitive aspects of gaze by jointly modeling attention, intentions, and tasks. We acknowledge the strong correlation between gaze and human trajectories and argue that gaze also significantly influences the next potential object to be interacted with. Therefore, we propose to simultaneously predict gaze, trajectories, and the next active object. \\

\noindent\textbf{Next Active Object Prediction.} Predicting the next active objects enables more granular user assistance \cite{plizzari2024outlook} and has rapidly gained attention \cite{furnari2017next, liu2020forecasting, liu2022joint, razali2022using, grauman2022ego4d, pasca2024summarize, ashutosh2024fiction}. Furnari \etal \cite{furnari2017next} pioneerly propose predicting which object will be interacted with next by training a classifier. The following works \cite{liu2020forecasting, liu2022joint, pasca2024summarize} try to address this problem by predicting the interaction hotspots on 2D egocentric videos. For example, Pasca \etal \cite{pasca2024summarize} leverage pre-trained image captioning and vision-language models to extract the action context, which is then processed to forecast the next object interaction. Besides, PickAndPlace \cite{razali2022using} introduces this problem to the robotic area and utilizes the human pose and gaze to predict the object to be picked and its place location. Ashutosh \etal \cite{ashutosh2024fiction} propose to anticipate all the subsequent interaction locations on the objects and corresponding human poses, given the observation of a person performing an activity. Although extensive research has been conducted on this problem, frameworks explicitly aligned with human cognition remain relatively underexplored. In contrast, we innovatively employ a cognition-aligned hierarchical framework to predict the next active objects and their future 3D trajectories, which achieves enhanced performance. 

%% file: sec/3_method.tex
\section{Method}
\subsection{Overview}

\noindent\textbf{Problem Formulation.} Given historical observations of a human in an environment with objects, we aim to predict where the human will go (\emph{trajectory}), where the human will look (\emph{gaze}), and which objects the human will interact with (\emph{object interaction}). 

Our method utilizes historical states such as gaze, head pose, and hand/body trajectories, which are readily available on modern AR/VR devices. In virtual environments, object interactions can be directly logged from the simulation, while in physical settings, they can be inferred via depth sensing or IoT-enabled objects embedded with RFID tags \cite{song2007proximity, zhou2009rfid} or inertial sensors \cite{kok2017using, shahbazi2023visual}.

Formally, we take $\frameHist$ historical frames as input, consisting of two components: (1) human states, including gaze directions as unit vectors $\gazeHist \in \R^{\frameHist \times 3}$, head orientations $\headRotHist \in \R^{\frameHist \times 3 \times 3}$, head trajectory $\headPosHist \in \R^{\frameHist \times 3}$, and hand trajectory $\handPosHist \in \R^{\frameHist \times 2 \times 3}$; (2) scene context, including object bounding boxes $\objBoxHist \in \R ^{\frameHist \times \Nobj \times 8 \times 3}$, object centers $\objCenterHist \in \R ^{\frameHist \times \Nobj \times 3}$, and semantic features $\objClipFeat \in \R ^ {\Nobj \times \dimClip}$ derived from object labels, where $\Nobj$ is the number of objects and $\dimClip$ is the semantic feature dimension. We predict the next $\framePred$ frames of human states, including gaze directions $\gazePred$, head orientations $\headRotPred$, head trajectory $\headPosPred$, and hand trajectory $\handPosPred$, as well as object interactions, including the trajectories of object centers $\objCenterPred$, and interaction states $\interactLabel \in \{0,1\}^{\Nobj}$, indicating whether human-object interactions occur.  \\

\noindent\textbf{Architecture.} As illustrated in \cref{fig:pipeline}, our framework consists of three core components: (1) an observation encoding module that processes historical human states and scene context into spatio-temporal features (\cref{subsection:encode}), (2) a hierarchical intention-aware decoder that first predicts potential interaction targets and forecasts specific next human states and object interactions (\cref{subsection:decode}), and (3) a context-aware dynamic \ac{gcn} that models human-object relationships and improves prediction accuracy (\cref{subsection:DGCN}). In the following, we will detail the three modules, respectively, and introduce the training losses in \cref{subsection:loss}.

\subsection{Observation Encoding} \label{subsection:encode}
Our observation encoding transforms historical human states ($\gazeHist$, $\headRotHist$, $\headPosHist$, $\handPosHist$) and scene context ($\objBoxHist$, $\objCenterHist$, $\objClipFeat$) into spatio-temporal representations. As shown in \cref{fig:pipeline}, we first apply \ac{dct} \cite{ma2022progressively, mao2020history} to convert all input time-series data to the frequency domain following practices in human motion prediction, and then process each input through specialized encoding networks.

To encode human state, gaze vectors $\gazeHist$ and head rotations $\headRotHist$ are processed through separate \ac{mlp} networks to produce latent embeddings $\fgaze$ and $\fheadrot$, both of size $\R^{\frameHist \times \D}$, where $\D$ represents the feature dimension. Head-hand trajectory $\headPosHist$ and $\handPosHist$ are encoded using an $\layere$-layer spatio-temporal \ac{gcn} that captures temporal dynamics and spatial relationships to get corresponding feature $\fheadpos \in \R^{\frameHist \times \D}$ and $\fhandpos \in \R^{\frameHist \times 2 \times \D}$. The \ac{gcn} incorporates a global node $\fglobal \in \R^{\D}$ to aggregate joint correspondence between hands and head across frames. Following \cite{hu2024hoimotion}, each \ac{gcn} component includes a temporal \ac{gcn} (T-GCN) and a spatial \ac{gcn} (S-GCN). The same \ac{gcn} structure is used in subsequent stages.

To encode scene context, we combine semantic priors with geometric information to represent the objects. Specifically, we process the object type labels through a pre-trained and frozen CLIP \cite{radford2021learning} encoder to extract semantic features. Object bounding box trajectory $\objBoxHist$ and object center trajectory $\objCenterHist$ are processed through MLPs to obtain spatial features. These semantic and spatial features are then fused by element-wise addition to produce object features $\fobj \in \R^{\frameHist \times \Nobj \times \D}$.

\subsection{Hierarchical Intention-Aware Decoding} \label{subsection:decode}
Drawing inspiration from cognitive processes where humans form interaction intentions before executing actions \cite{pacherie2008phenomenology}, we design our decoding process with a hierarchical structure. We first introduce an interaction intention prediction module (\cref{fig:pipeline}) to process the encoded historical human gaze patterns, trajectory cues, and scene object properties to predict the interaction probabilities of each object to select potential interaction targets. Then, we decode the next states of human and objects by integrating selected object features with updated human state features.

Specifically, the interaction intention prediction module processes three encoded streams: global pose features $\fglobal$ encoding positional trajectory dynamics, gaze features $\fgaze$ reflecting attentional focus, and object features $\fobj$ capturing object semantics and spatial relationships. An $\layeri$-layer \ac{gcn} encoder $\fgcn$ fuses the information and updates them to $\fglobalupdate$, $\fgazeupdate$, and $\fobjupdate$. The updated object features are then mapped to the interaction probabilities $\interactLabelRawPred \in [0, 1]^{\Nobj}$ via an \ac{mlp} $\fmlpselect$, where $\hat{}$ denotes the predicted variables:
\begin{equation}
\begin{split}
\fglobalupdate, \fgazeupdate, \fobjupdate &= \fgcn([\fglobal, \fgaze, \fobj]), \\
\interactLabelRawPred &= \phi(\fmlpselect(\fobjupdate)),
\end{split}
\end{equation}
where $[\cdot]$ denotes concatenation, and $\phi$ denotes the sigmoid function, mapping the \ac{gcn} output to a probability distribution in range $[0, 1]$. Based on the interaction probabilities $\interactLabelRawPred$, the interaction intention prediction module selects the top $\topk$ object features $\ftopk \in \R^{\frameHist \times \topk \times \D}$ from $\fobjupdate$, keeping only the target candidates with the highest interaction likelihood. 

The selected object features $\ftopk$, together with the updated human state features including $\fgazeupdate$, $\fglobalupdate$, $\fheadpos$, $\fhandpos$ and $\fheadrot$, are then passed to an $\layerd$-layer spatio-temporal \ac{gcn} decoder for next state decoding. Finally, a prediction network that integrates \ac{gcn}, \ac{mlp} and \ac{idct} decodes the detailed human-object states, including next human states with gaze $\gazePred$, head orientation $\headRotPred$, head trajectory $\headPosPred$ and hand trajectory $\handPosPred$, as well as object interactions including object center trajectory $\objCenterPred$ and refined interaction states $\interactLabelPred \in [0,1]^{\topk}$.

\subsection{Dynamic GCN for Adaptive Human-Environment Modeling} \label{subsection:DGCN}

To better model the human-environment relationships in the interaction intention prediction module, we design a \ac{dgcn} to model the relationship among the gaze $\fgaze$, the joint human head and hand feature $\fglobal$, and the $\Nobj$ objects $\fobj$, using an adaptive weight matrix $\AdjD \in \R ^ {(\Nobj+2) \times (\Nobj +2)}$. Specifically, our adaptive weight matrix $\AdjD$ incorporates four components: gaze-object weights $\Adjg \in \R ^ {\Nobj}$, position-object weights $\Adjp \in \R ^ {\Nobj}$, object-object weights $\Adjo \in \R ^ {\Nobj \times \Nobj}$, and a fixed gaze-position weight of 1. See \cref{fig:pipeline} for illustration.

To obtain the gaze-object weights $\Adjg$, we first calculate the angle $\anglegaze$ between the gaze direction $\gazeHist$ and the lines connecting the head position to the $\Nobj$ object center position $\distgaze = \objCenterHist - \headPosHist$\footnote{In implementation, we repeat the head position $\Nobj$ times to accomplish this operation but omit the details for clarity.}, according to \cref{eq:angle}. Then these features are fed into an \ac{mlp} $\fmlpgaze$ to predict the weight $\Adjg$ as in \cref{eq:Adjg}:
\begin{equation}
\label{eq:angle}
    \anglegaze = \cos^{-1}(\frac{\gazeHist \cdot \distgaze}{\|\gazeHist\| \|\distgaze\|}),
\end{equation}
\begin{equation}
\label{eq:Adjg}
    \Adjg = \fmlpgaze([\anglegaze, \|\distgaze\|]),
\end{equation}
where $\anglegaze \in \R^{\frameHist \times \Nobj}$, and $\|\distgaze\| \in \R^{\frameHist \times \Nobj}$ is the Euclidean distance between head and $\Nobj$ objects.

For position-object weights $\Adjp$, we first calculate the minimum Euclidean distance $\distpose \in \R^{\Nobj}$ between each object center and the three key positions: the head $\headPosHist$, the left hand $\handPosHist^0$, and the right hand $\handPosHist^1$, averaged across $\frameHist$ frames. Formally, the distance for $i$-th object $\distpose^i$ and the corresponding position-object weights $\Adjp^i$ are defined as:
\begin{equation}
\begin{split}
\distpose^i = \frac{1}{\frameHist} min (  &\sum_{t=1}^{\frameHist}\|\objCenterHist^{i,t} - \headPosHist^t\|, \sum_{t=1}^{\frameHist}\|\objCenterHist^{i,t} - \handPosHist^{0,t}\|,  \\ &\sum_{t=1}^{\frameHist}\|\objCenterHist^{i,t} - \handPosHist^{1,t}\| ),
\end{split}
\end{equation}
\begin{equation}
\Adjp^i = \exp(-\distpose^i),
\end{equation}
where $\headPosHist^t$ represents the head position at frame $t$, and $\handPosHist^{0,t}$ and $\handPosHist^{1,t}$ represent the positions of the left and right hands at frame $t$.

For object-object weights $\Adjo$, we first compute the average Euclidean distance $\distobj \in \R^{\Nobj \times \Nobj}$ between each two objects across $\frameHist$ time frames and obtain the corresponding weight $\Adjo$ as in \cref{eq:Adjo}:
\begin{equation}
\distobj^{i,j} = \frac{1}{\frameHist} \sum_{t=1}^{\frameHist} \|\objCenterHist^{i,t} - \objCenterHist^{j,t}\|,
\end{equation}
\begin{equation}
\label{eq:Adjo}
\Adjo^{i,j} = \exp(-\distobj^{i,j}),
\end{equation}
where $\objCenterHist^{i,t}$ and $\objCenterHist^{j,t}$ represent the center positions of the $i$-th and $j$-th objects at frame $t$.

These weight components are integrated into the dynamic weight matrix $\AdjD$ to enhance the node feature update process in the spatial \ac{gcn} (S-GCN):
\begin{equation}
\NodeFeat^{(l+1)} = \AdjNorm^{-1/2} \AdjD \AdjNorm^{-1/2} \NodeFeat^{(l)} \WeightMat^{(l)},
\end{equation}
where $\NodeFeat^{(l)} = [\fglobal^{(l)}, \fgaze^{(l)}, \fobj^{(l)}]$ are the node features at \ac{gcn} layer $l$, $\AdjNorm$ is the degree matrix of $\AdjD$ and $\WeightMat^{(l)}$ are learnable parameters for the $l$-th layer.

\subsection{Training Losses} \label{subsection:loss}
Our training objectives directly supervise all predicted outputs. Each component of the loss corresponds to a specific prediction head in our framework: 
\begin{equation}
\begin{split}
\loss ={} & \lambda_{\text{gaze}}\lossgaze + \lambda_{\text{rot}}\lossrot + \lambda_{\text{pos}}\losspos + \lambda_{\text{center}}\lossobjpos \\
+ &\lambda_{\text{int}}\lossselect + \lambda_{\text{state}}\losslabel + \lambda_{\text{vel}}\lossvel,
\end{split}
\label{eq:total_loss_multiline}
\end{equation}
where $\lossgaze$ measures the error between predicted and \ac{gt} gaze directions using cosine similarity; $\lossrot$ is L2 loss of head orientation; $\losspos$ is L2 losses of head, hands trajectory positions, and $\lossobjpos$ is L2 loss of object center trajectories; $\lossselect$ and $\losslabel$ use binary cross-entropy loss between predicted interaction probabilities and \ac{gt} labels in the candidate selection and decoding stages, respectively. $\lossvel$ promotes smooth human and object trajectories by calculating L2 distance between predicted and \ac{gt} velocities; The hyperparameters $\lambda_{*}$ balance the contributions of these loss terms. Please refer to the supplementary material for definitions of these terms.

%% file: sec/4_experiment.tex
\section{Experiment}
\subsection{Setup}
\begin{figure}[t]
  \centering
  \includegraphics[width=0.45\textwidth]{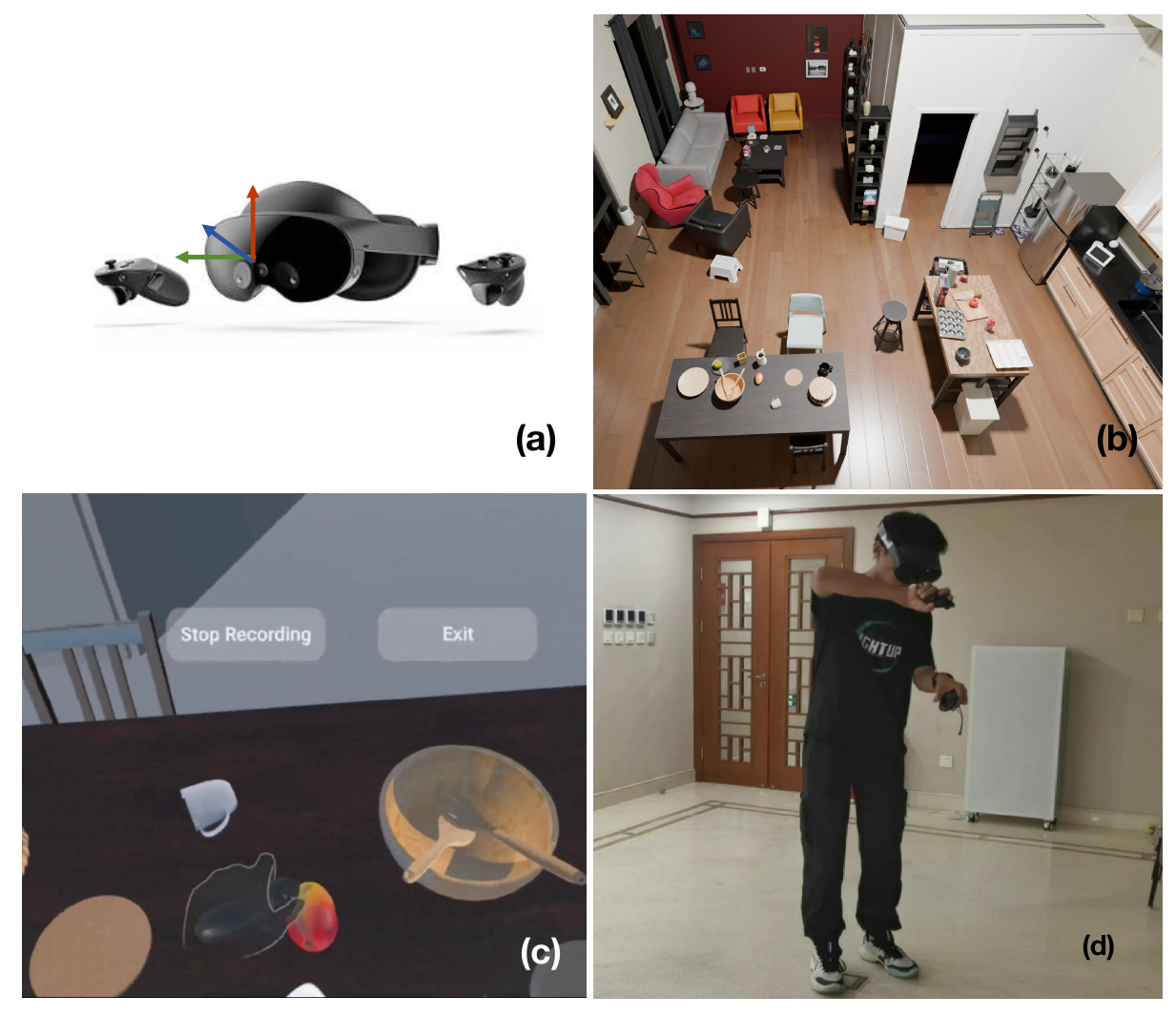} 
\caption{
\textbf{Experimental setup for the real-world VR evaluation.} (a) Meta Quest Pro headset and Touch controllers used for interaction and data collection. (b) A top-down view of the interactive virtual environment.
(c) The participant's egocentric view of the interactive VR scene. (d) Third-person view of a participant in the real world performing a natural interaction sequence with the virtual environment.
}
  \label{fig:setup}
\end{figure}

\begin{table*}[t] 
\centering 
\setlength{\tabcolsep}{1.3mm}
\resizebox{\linewidth}{!}{
\begin{tabular}{l |cccccc | cc} 
\thickhline
\multirow{2}{*}{Method}  & \multicolumn{6}{c | }{ADT} & \multicolumn{2}{ c}{ADT-Hard} \\
\cline{2-7} \cline{8-9} 
& Hand $\downarrow$ & Head (Dist.) $\downarrow$ & Head (Dir.) $\downarrow$ & Gaze $\downarrow$ & Object Center $\downarrow$ & Object AP $\uparrow$ & Object Center $\downarrow$ & Object AP $\uparrow$ \\
\hline
Pose2Gaze \cite{hu2024pose2gaze} & -&- & -& 16.55 & -&- & -&- \\
SIF3D \cite{lou2024multimodal} & 102.70 & 63.92 & -& -& -& -& -&- \\
HOIMotion \cite{hu2024hoimotion} & 79.58 & 45.51 & -& -& -& -& -&- \\
GazeMotion \cite{hu24gazemotion} & 169.30  & 133.10 & - & 14.82 & -&  -& -&- \\
PickAndPlace \cite{razali2022using} & 100.82 & 61.98 & - & -& 113.80 & 68.72 & 85.90 & 35.63  \\
\textbf{Ours} & \textbf{78.77} & \textbf{44.15} & \textbf{8.10} & \textbf{14.02} & \textbf{88.02} & \textbf{98.56} & \textbf{84.78} & \textbf{70.49} \\
\thickhline
\end{tabular}}  
\caption{\textbf{Comparison to the state-of-the-arts on ADT \cite{pan2023aria} dataset and ADT-Hard subset.} Missing entries (-) indicate that the methods do not predict the corresponding properties.}
\label{tab:sota} 
\end{table*}

\noindent\textbf{Dataset.} 
We train our method and conduct experiments on the Aria Digital Twin (ADT) dataset \cite{pan2023aria}, captured by human participants wearing Aria glasses while performing real-world activities in two realistic indoor scenes. The dataset includes 236 sequences, 2 real indoor scenes, and 398 object geometry models. The sequences, recorded at 30 Hz, contain multi-modal data such as egocentric RGB videos, 6DoF head device poses, 3D eye gaze data, \etc. We use 70 sequences with hand positions, splitting them randomly into 60 for training and 10 for testing. 
Additionally, we curate a specialized subset called ``ADT-Hard'', consisting of clips where object interaction states change between input and prediction periods, testing models' ability to anticipate new interactions beyond ongoing ones. We conduct all experiments on this subset as well.  

For our real-world practical evaluation, we collect 6 sequences with an average duration of 1 minute from human participants who are instructed to perform natural interaction sequences in a VR environment (see \cref{fig:setup} (d)). We apply no preprocessing to the data to preserve authentic characteristics like sensor noise and behavioral variations, enabling a realistic test of our model's robustness. 

Our method is trained on publicly available datasets and does not involve the use of facial or other privacy-sensitive biometric information. the data we used is obtained with the consent of the participants.\\

\noindent\textbf{Real-world VR Experimental Setup.}
To evaluate the practical applicability of our approach, we conduct user studies in a VR environment with human participants. We use a Meta Quest Pro headset and Meta Touch controllers for the system, as shown in \cref{fig:setup}(a). The headset displays the VR interface and captures gaze vectors and 6DoF head pose, while the controllers track hand trajectories and enable interaction with virtual objects. The real-time positions and states of all virtual objects are recorded within the Unity scene during interactions.

We construct a virtual environment using object assets from the ADT dataset within the Unity Engine (v6.1), creating an interactive room scenario that mirrors real-world domestic environments, as shown in \cref{fig:setup}(b) and (c). The scene is rendered and displayed to users in a Meta Quest Pro headset. \\

\noindent\textbf{Baselines.} Since no directly comparable baseline exists, we evaluate our method against five most recent related \ac{sota} approaches, including Pose2Gaze \cite{hu2024pose2gaze}, SIF3D \cite{lou2024multimodal}, HOIMotion \cite{hu2024hoimotion}, GazeMotion \cite{hu24gazemotion}, and PickAndPlace \cite{razali2022using}. We use their official code for training and evaluating on the ADT dataset. Please refer to supplementary material for more implementation details. \\

\noindent\textbf{Metrics.} Follow common practice \cite{razali2022using,lou2024multimodal, bao2023uncertainty, hu24gazemotion, hu2024pose2gaze, hu2024hoimotion}, we evaluate the approaches using the standard metrics: (1) For trajectory forecasting, we report the mean L2 distance error (in millimeters) between predicted and \ac{gt} positions for Hand (mm), Head (mm), and Object Center (mm); (2) For orientation prediction, we measure the average angular error in degrees for Head (°) and Gaze (°); (3) For object interaction prediction, we compute Average Precision (AP) by calculating the area under the Precision-Recall curve, varying the interaction threshold and aggregating precision-recall pairs. \\

\begin{figure*}
  \centering
  \includegraphics[width=0.86\textwidth]{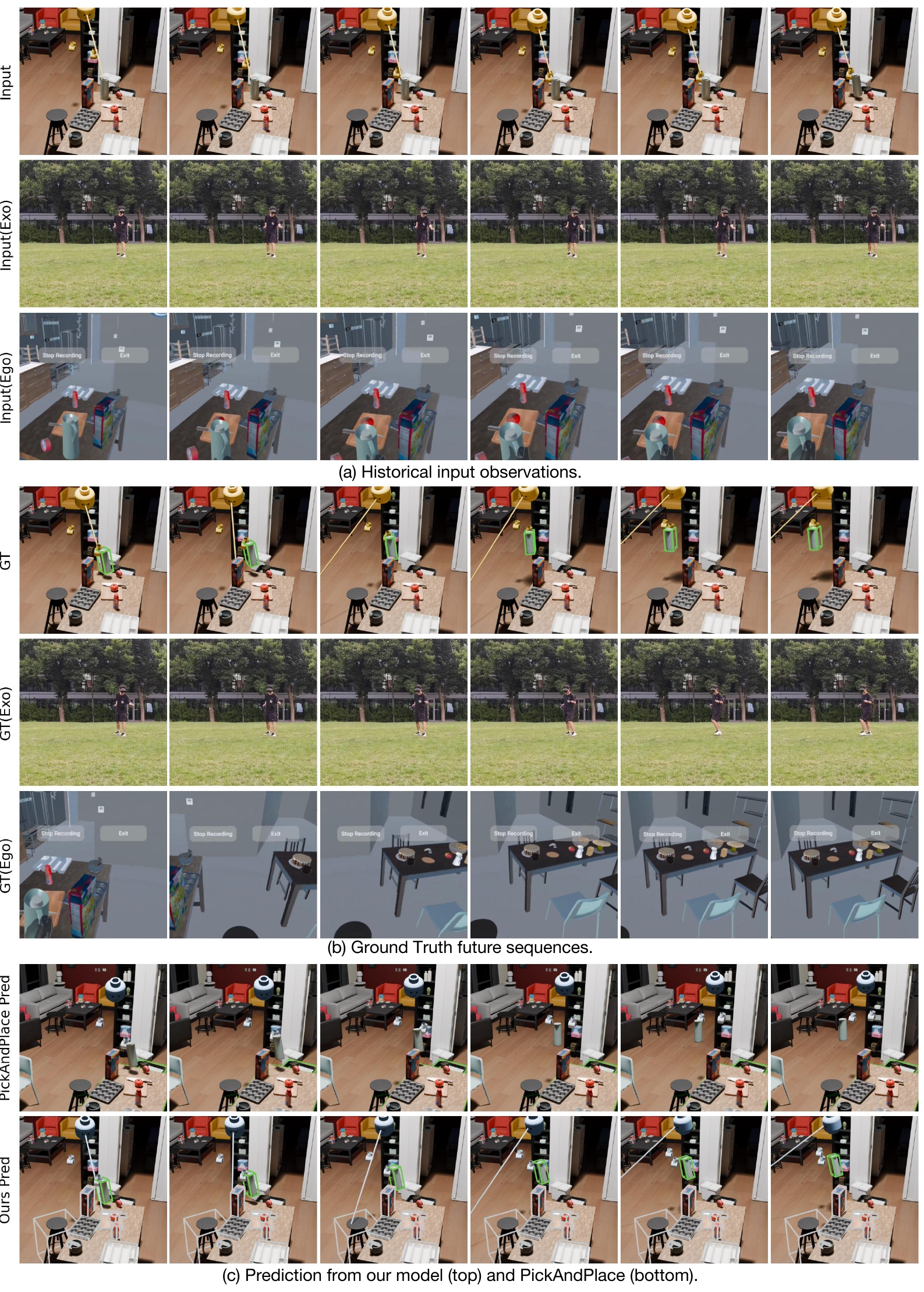} 
  \vspace{-8pt}
      \caption{
        \textbf{Comparison of our method with PickAndPlace~\cite{razali2022using} in a real-world VR environment.} The columns in each subfigure represent consecutive timesteps. Input sequences (a) and ground truth (b) are shown across three forms: rendered visualization, exocentric view, and egocentric view. (c) presents visualization of predictions from PickAndPlace (top) and our method (bottom).
        Visual elements include gaze direction (rays), head pose, hand positions (Lego figures), and interaction targets (bounding boxes). Bold bounding boxes indicate the top-$\topk$ interaction candidates from our hierarchical framework, while the white-to-green gradient represents increasing predicted interaction probabilities.
    }
    \label{fig:vr}
    \vspace{-10pt}
\end{figure*}

\begin{figure}[h]
  \centering
  \includegraphics[width=0.48\textwidth]{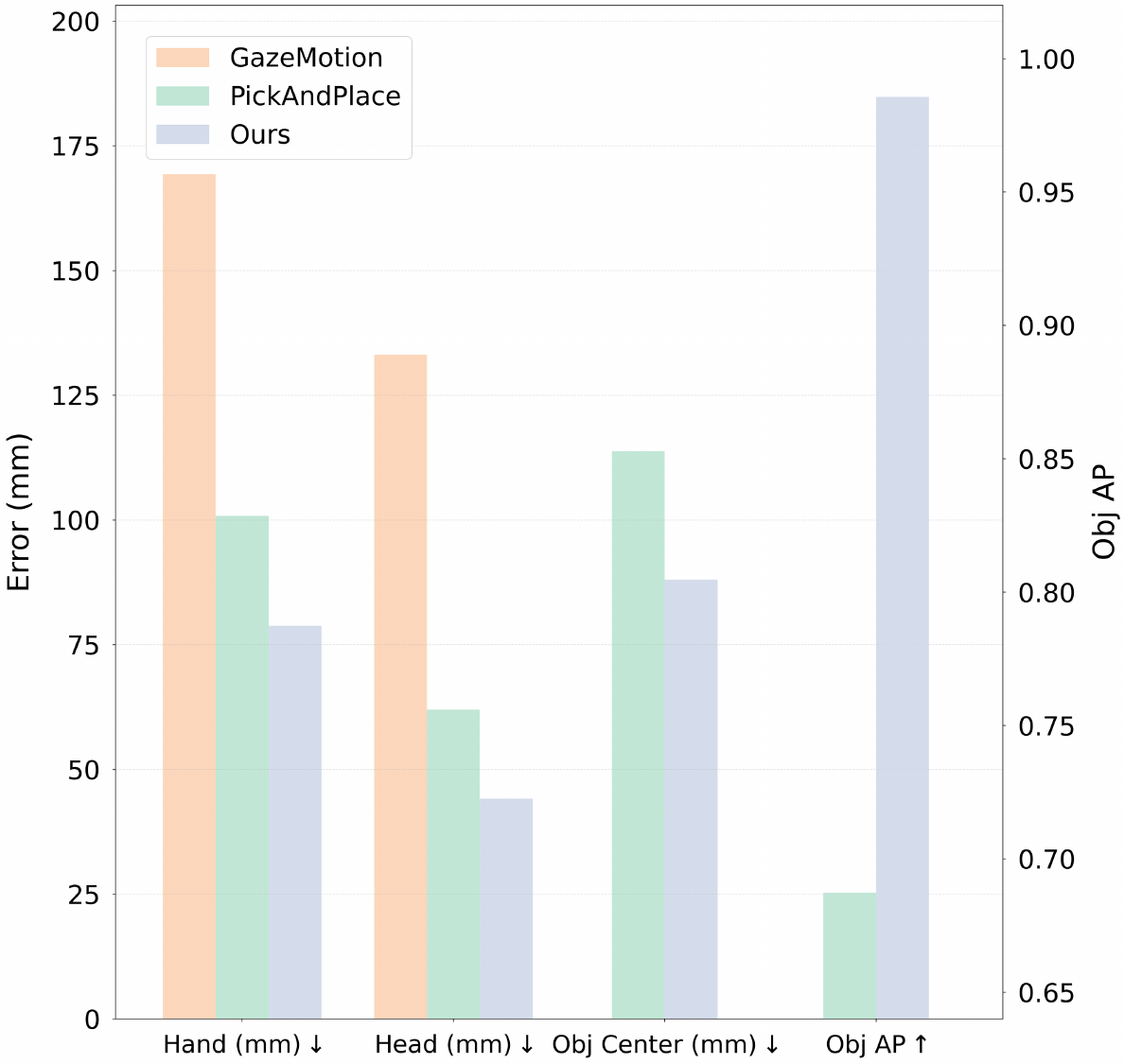} 
\caption{
\textbf{Quantitative comparison of our method with PickAndPlace \cite{razali2022using} and GazeMotion \cite{hu24gazemotion} on real-world VR evaluation.} We show errors on hand, head, and object center prediction as well as object interaction AP.
}
  \label{fig:vrresult}
 \vspace{-0.5cm}
\end{figure}

\begin{figure*}[htbp] 
  \centering
  \includegraphics[width=0.95\textwidth]{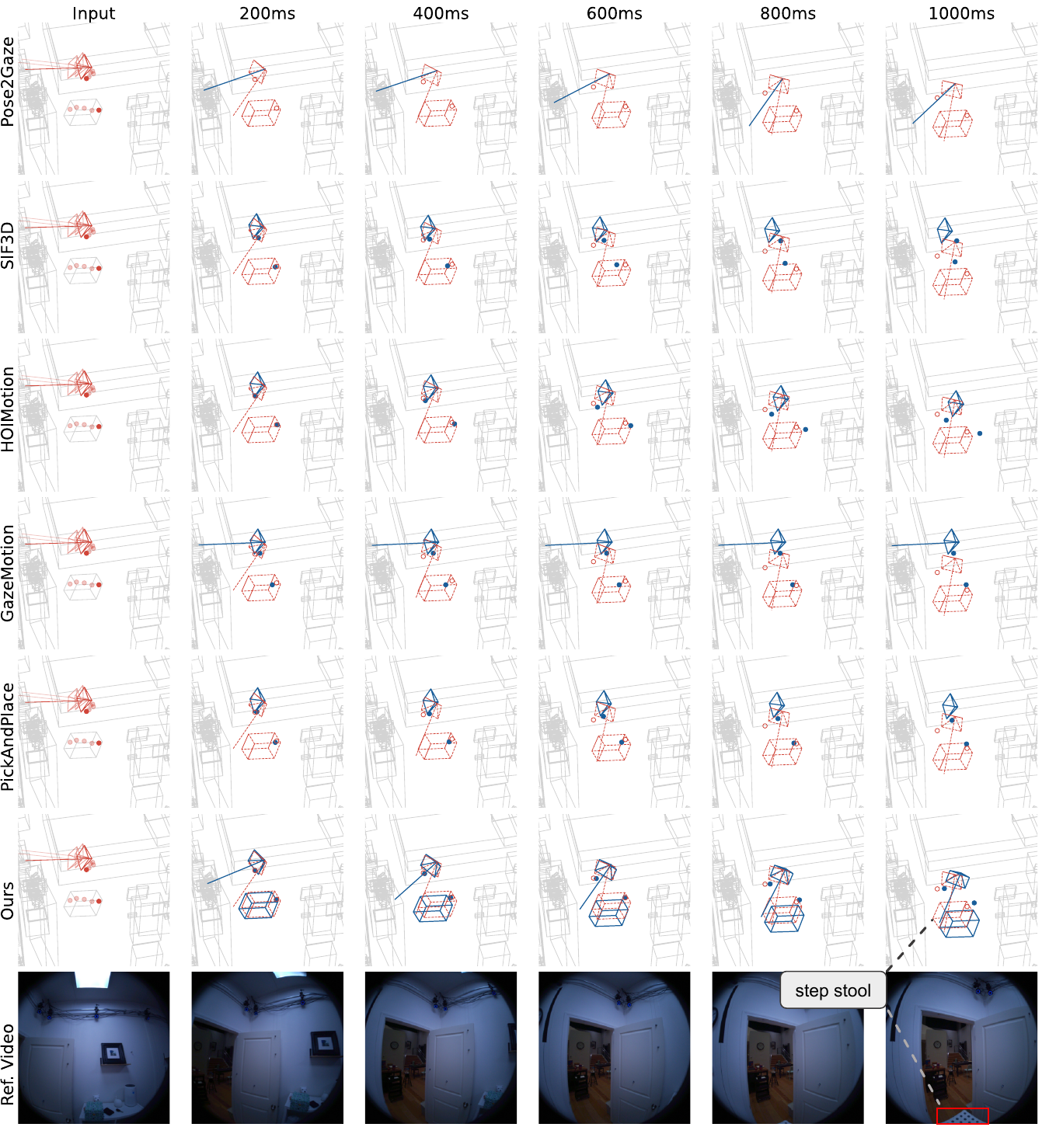} 
  \caption{\textbf{Qualitative comparison of our method with state-of-the-art methods on ADT \cite{pan2023aria} dataset.} The leftmost column shows the historical input observation, where color shades from light to dark represent progression from earlier to later time, while the remaining columns display predictions at 200ms intervals from 200ms to 1000ms. In each visualization, red and blue elements represent \ac{gt} and predictions, respectively. The visual elements include: gaze direction (ray), head position and orientation (pyramid), hand positions (two points), and interacted objects (bounding boxes). The interaction object type is labeled in the final frame. The bottom row shows reference frames from the first-person perspective video.} 
  \label{fig:sota} 
\end{figure*}

\noindent\textbf{Implementation Details.}
Our model takes historical $\frameHist=15$ frames as input and predicts the future $\framePred=15$ frames, by sampling 30 Hz data with stride 2 to process 1 second of history for 1 second of future prediction. We encode $\Nobj=48$ objects, with our hierarchical decoder selecting the top $\topk=12$ objects. Our model has $\layere = 4$ encoding \ac{gcn} layers, $\layeri = 4$ \ac{dgcn} layers, and $\layerd = 16$ decoding layers with $\D = 16$ feature dimension. Our model achieves real-time inference at 33 FPS on a single NVIDIA 2080Ti GPU. We provide more details in \cref{supp:detail}. \\

\subsection{Real-world Experiments in a VR Environment}
\label{sec:vrdemo}

\begin{table*}[t!] 
\centering 
\setlength{\tabcolsep}{1mm}
\resizebox{\linewidth}{!}{
\begin{tabular}{l | cccccc | cc} 
\thickhline
\multirow{2}{*}{Method}  & \multicolumn{6}{c|}{ADT} & \multicolumn{2}{c}{ADT-Hard} \\
\cline{2-7} \cline{8-9} 
& Hand $\downarrow$ & Head (Dist.) $\downarrow$ & Head (Dir.) $\downarrow$  & Gaze $\downarrow$ & Object Center $\downarrow$ & Object AP $\uparrow$ & Object Center $\downarrow$ & Object AP $\uparrow$ \\
\hline
(a) \textit{w/o} Hier. Decoding & 81.16 & 45.10 & \textbf{8.08} & 14.05 & 97.19 & 97.42 & 97.97 & 60.84 \\
(b) \textit{w/o} Dynamic GCN & 79.63 & 44.65 & 8.12 & 14.08 & 89.37 & 98.01 & 85.64 & 66.70 \\
(c) \textbf{Ours} & \textbf{78.77} & \textbf{44.15} & 8.10 & \textbf{14.02} & \textbf{88.02} & \textbf{98.56} & \textbf{84.78} & \textbf{70.49} \\
\thickhline
\end{tabular}}
\caption{\textbf{Ablation study of our approach on ADT \cite{pan2023aria} dataset and ADT-Hard subset.} Ablation (a) removes the interaction intention prediction module (including the dynamic GCN) and directly decodes the next human states and object interactions. Ablation (b) replaces the dynamic GCN in the intention prediction module with vanilla GCN by removing the dynamic weight matrix $\AdjD$.}
\label{tab:ablation} 
\end{table*}

\begin{figure*}[h!] 
  \centering
  \includegraphics[width=1\textwidth]{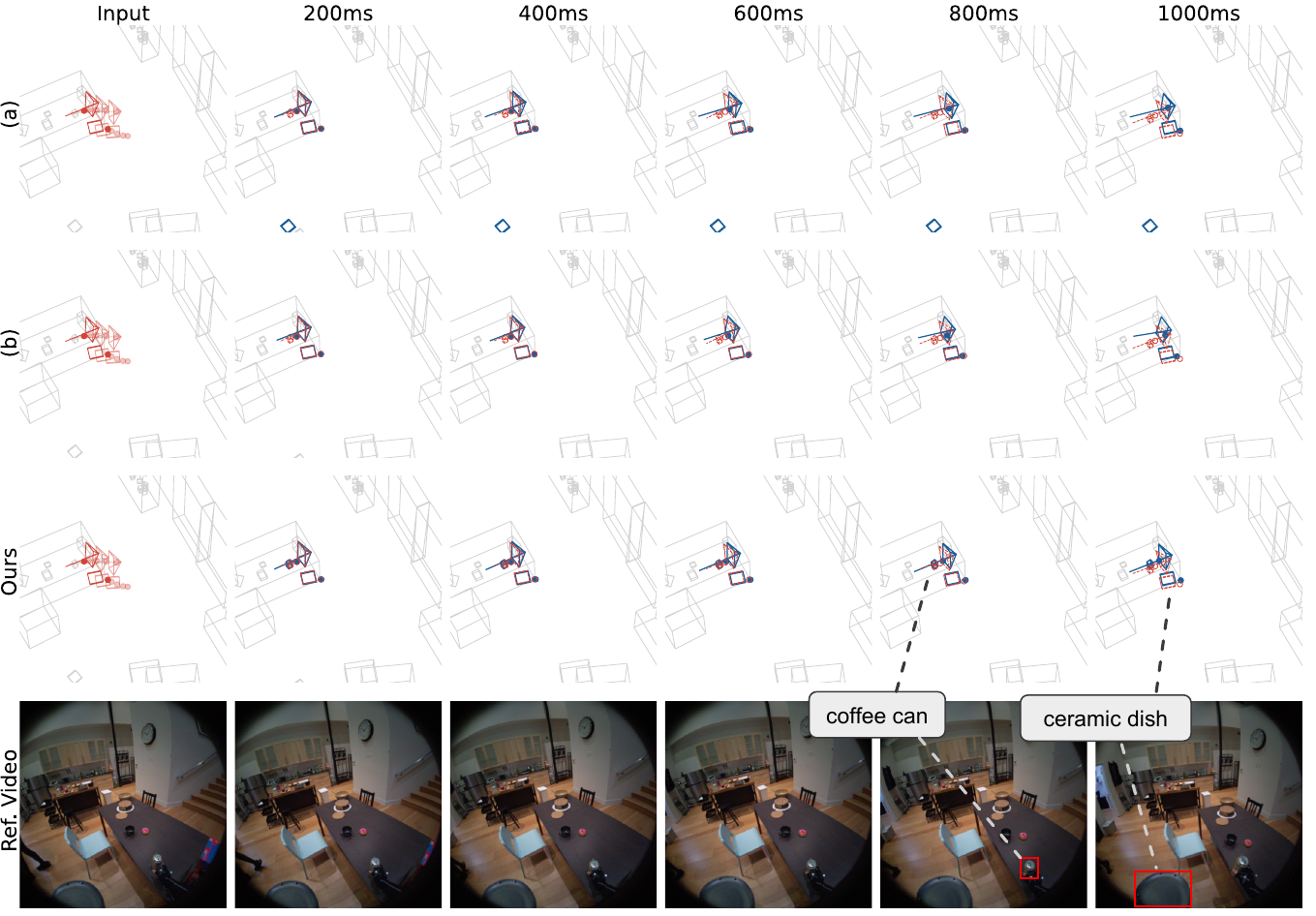} 
  \caption{\textbf{Qualitative comparison of our full model with ablated baselines on ADT \cite{pan2023aria} dataset.} Ablation (a) removes the interaction intention prediction module and directly decodes the next human states and object interactions. Ablation (b) replaces the dynamic GCN with vanilla GCN by removing the dynamic weight matrix $\AdjD$. The visualization format is kept the same as in \cref{fig:sota}.} 
  \label{fig:ablation} 
\end{figure*}

\noindent\textbf{Quantitative Results.} \cref{fig:vrresult} presents the quantitative performance of our method and the most related baseline PickAndPlace \cite{razali2022using} and GazeMotion \cite{hu24gazemotion} on real-world VR interaction sequences. Driven by our cognition-inspired design that hierarchically models human-object interaction intentions, our approach exhibits robust performance under noisy realistic conditions and significantly outperforms PickAndPlace \cite{razali2022using} and GazeMotion \cite{hu24gazemotion} across the metrics. Notably, our method achieves a substantially higher Object Average Precision, which highlights its strength in predicting users’ upcoming interaction targets. This capability is indispensable for proactive VR applications, such as intention-aware interaction assistance. \\

\noindent{\textbf{Qualitative Results.}}
A qualitative result from a captured live interaction sequence is presented in \cref{fig:vr}, which demonstrates our method's superior predictive capabilities compared to PickAndPlace \cite{razali2022using}. Our model successfully anticipates user gaze patterns, future pick-up behaviors, and upcoming interaction objects (a kettle), while PickAndPlace produces incorrect predictions. 

Notably, This visualization provides clear evidence for the effectiveness of our cognition-aligned hierarchical framework. As demonstrated by the bold bounding boxes, the framework first generates contextually coherent candidates that are semantically and spatially relevant, providing meaningful contextual information for subsequent detailed prediction. Furthermore, as revealed by the bounding box color gradients, our final interaction probability predictions accurately reflect true user intentions. 
The intended target is highlighted in dark green, corresponding to a high interaction probability, while other plausible candidates remain white due to their low assigned probabilities. These results demonstrate that our method can produce accurate and reasonable predictions that can potentially support improved downstream applications such as personal assistants and adaptive game interactions. For more visualizations, please refer to our video demonstration.

\subsection{Experiments on ADT dataset}
\label{sec:sota}

While our user study confirmed the framework's real-world feasibility and provided qualitative insights into its usability, a systematic and quantitative comparison is essential to validate its algorithmic superiority. Therefore, we compare our method to state-of-the-art methods on the ADT dataset \cite{pan2023aria} and its more challenging subset ADT-Hard. 

As shown in \cref{tab:sota}, GazeMotion \cite{hu24gazemotion} and Pose2Gaze \cite{hu2024pose2gaze} do not incorporate environmental context, which limits their performance. SIF3D \cite{lou2024multimodal}, HOIMotion \cite{hu2024hoimotion}, and PickAndPlace \cite{razali2022using} utilize environmental information, such as point clouds and object bounding boxes, but lack modeling of the human-environment relationship and interaction intentions,  lead to reduced prediction accuracy. 
Our model achieves superior performance across all metrics and demonstrates a significant advantage in recognizing the next active objects on the ADT-Hard dataset, validating the robustness of our proposed cognition-aligned hierarchical decoding and dynamic weight mechanism.

\cref{fig:sota} shows a qualitative comparison result with the baselines. Notably, even when the input gaze does not fixate on the target object, our method still accurately predicts the next active object and generates contextually appropriate situated human behavior. In contrast, PickAndPlace \cite{razali2022using} fails to identify the interaction target and generates inaccurate interaction behavior. Other baselines \cite{hu24gazemotion,hu2024hoimotion,lou2024multimodal,hu2024pose2gaze} do not explicitly model the human-object interaction intention. As a result, their predictions are limited to extending historical trajectories, failing to accurately identify upcoming interaction targets or predict correct human states, leading to noticeable discrepancies from the \ac{gt}. Please refer to supplementary material for additional qualitative comparison and video results.

\subsection{Ablation Study}
\begin{figure}[t!]
  \centering
  \includegraphics[width=0.5\textwidth]{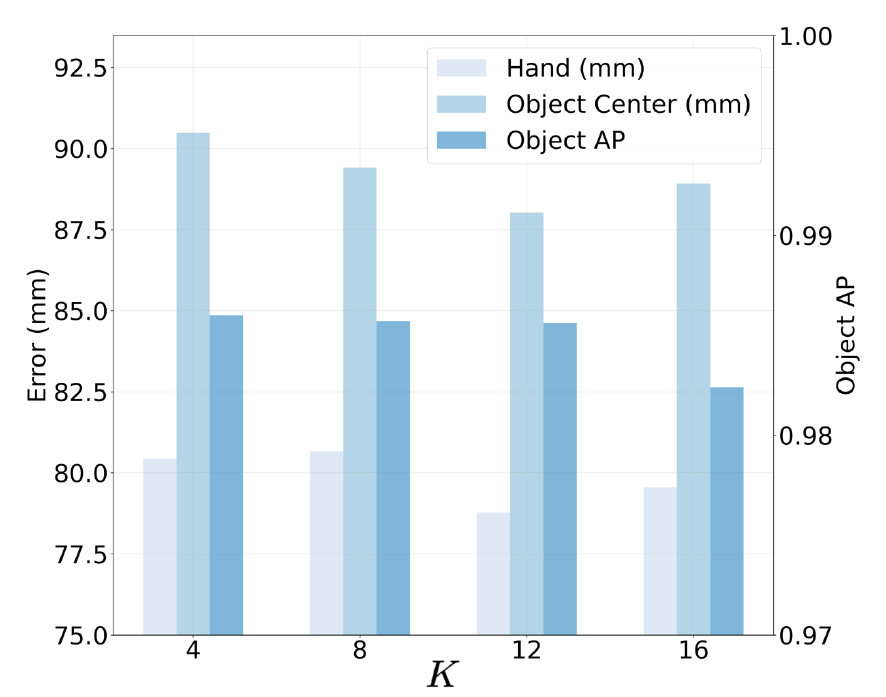}
  \caption{\textbf{Ablation study on different numbers of selected objects ($\topk$).} We show errors on hand and object center prediction as well as the object interaction AP on the ADT dataset.} 
  \label{fig:topk}
\end{figure}
\noindent\textbf{Ablation of Each Module.} To validate the effectiveness of the key components in our framework, we compare our approach to two ablations in \cref{tab:ablation} on the ADT \cite{pan2023aria} dataset and ADT-Hard subset. In ablation (a), we remove the entire hierarchical decoding structure, including the intention prediction module and dynamic GCN and directly decoding human states and object interactions with a vanilla GCN decoder of the same number of layers. In ablation (b), we replace the dynamic \ac{gcn} with vanilla \ac{gcn} by removing the dynamic weight matrix $\AdjD$.

As shown in \cref{tab:ablation}, removing either module significantly degrades performance across most metrics. (a) The removal of hierarchical decoding leads to substantial performance drops in object prediction and AP scores, demonstrating that our approach of predicting intentions before detailed decoding is critical for accurate interaction prediction in complex scenarios. Without the (b) dynamic weight matrix, we observe increased errors in position predictions and decreased AP scores, confirming that our context-aware mechanism effectively improves human-environment relationship modeling.

\cref{fig:ablation} presents a representative qualitative comparison example from the ADT-Hard subset, illustrating the contribution of each module to accurate predictions. Our full model successfully predicts both the ceramic dish and coffee can interactions. In contrast, the model without (a) hierarchical decoding not only misses the coffee can interaction but also predicts false positives, as seen in the blue bounding box in row (a) of \cref{fig:ablation}. Similarly, the model without (b) the dynamic GCN only predicts the ongoing dish interaction, but misses the new coffee can interaction. These differences become more pronounced in later predictions, such as at 1000ms (rightmost column). This demonstrates both components are essential for accurate interaction modeling in complex environments. Additional results are provided in the supplementary material.  \\

\noindent\textbf{Number of Selected Objects.}
We investigate the impact of the number of selected objects $\topk$ in our hierarchical intention-aware decoding framework on the ADT dataset. \cref{fig:topk} visualizes three
key metrics for human-object interaction performance: hand position error (mm), object center error (mm), and object interaction AP. We evaluate our method with $\topk$ ranging from 4 to 16 objects and find robust performance across all configurations.

As shown in \cref{fig:topk}, as $\topk$ increases from 4 to 12, hand position and object center errors decrease, with a slight trade-off in object interaction AP. At $\topk=16$, performance degrades, suggesting that while non-interacted objects with high interaction probability provide useful context, but excessive inclusion introduces redundancy that harms performance. We select $\topk=12$ for our final model, balancing performance across metrics.

%% file: sec/5_conclusion.tex
\section{Conclusion}
\label{sec:conclusion}

In this work, we present a cognition-inspired framework for predicting comprehensive situated human behavior, including trajectory, gaze, and object interaction. Drawing from cognitive science research, our hierarchical intention-aware decoding approach first predicts potential interaction targets before forecasting detailed human states and object interactions, mirroring how humans form intentions prior to executing actions. We develop a dynamic GCN with an adaptive weight matrix that effectively captures context-aware human-environment relationships. Extensive experiments on challenging real-world benchmark demonstrate that our approach significantly outperforms state-of-the-art methods, particularly in complex interaction scenarios where accurate intention prediction is crucial, validating our cognitively-aligned approach to situated human behavior prediction.  \\

\noindent\textbf{Limitations.} While the model occasionally makes incorrect predictions, they are often reasonable due to the diversity of human behaviors. This limitation stems from the current approach’s focus on a single deterministic prediction. Future work could explore modeling behavioral diversity to improve accuracy and robustness.

%% file: sec/X_suppl.tex
\newpage
\appendix
\setcounter{page}{1}
We provide additional method descriptions and experimental setup in this supplementary material.

\section{Method Details}
\label{supp:sec:method}
\subsection{Detailed Training Losses}
\label{supp:loss}
Our training objectives consist of multiple components:
\begin{equation}
\begin{split}
\loss = & \lambda_{\text{gaze}}\lossgaze + \lambda_{\text{rot}}\lossrot + \lambda_{\text{pos}}\losspos + \lambda_{\text{center}}\lossobjpos \\
 + & \lambda_{\text{vel}}\lossvel + \lambda_{\text{int}}\lossselect + \lambda_{\text{state}}\losslabel,
\end{split}
\end{equation}

$\lossgaze$ is the gaze direction loss using cosine similarity to align predicted gaze direction $\gazePred$ with ground truth $\gazeGT$, where $\Tilde{}$ denotes the ground truth:
\begin{equation}
    \lossgaze = 1 - \frac{\gazePred \cdot \gazeGT}{\|\gazePred\| \|\gazeGT\|}.
\end{equation}

$\lossrot$ measures head rotation accuracy using mean squared error between predicted and ground truth head rotations:
\begin{equation}
    \lossrot = \| \headRotPred - \headRotGT \|^{2}.
\end{equation}

$\losspos$ computes position error by summing L2 distances between predicted and ground truth trajectories for both head and hands:
\begin{equation}
    \losspos = \| \headPosPred - \headPosGT \|_{2} +  \| \handPosPred - \handPosGT \|_{2}.
\end{equation}

$\lossobjpos$ measures object center trajectory accuracy using L2 distance between predicted and ground truth object centers:
\begin{equation}
    \lossobjpos = \|\objCenterPred - \objCenterGT\|_{2}.
\end{equation}

$\lossvel$ ensures trajectory smoothness by calculating L2 distances between predicted and ground truth velocities for head, hands, and object centers:
\begin{equation}
\begin{split}
    \lossvel = &\| \headVelPred - \headVelGT \|_2 + \| \handVelPred - \handVelGT \|_2 \\ + &\| \objCenterVelPred - \objCenterVelGT \|_2,
\end{split}
\end{equation}
where $\headVelPred$ and $\headVelGT$ are the predicted and ground truth head velocities, $\handVelPred$ and $\handVelGT$ are the predicted and ground truth hand velocities, and $\objCenterVelPred$ and $\objCenterVelGT$ are the predicted and ground truth object center velocities, respectively. For all components, velocities are computed as the difference between consecutive positions. For example, the ground truth head velocity at frame $i$ is calculated as:
\begin{equation}
    \headVelGT^{i} = \headPosGT^{i+1} - \headPosGT^{i}.
\end{equation}
The velocities for hands and object centers follow the same computation pattern using their respective position variables.

$\lossselect$ uses binary cross-entropy to supervise the predicted interaction probability $\interactLabelRawPred$ from the Interaction Target Prediction module against the ground truth interaction states $\interactLabelGT$:
\begin{equation}
    \lossselect = \interactLabelGT \cdot \log(\interactLabelRawPred) + (1 - \interactLabelGT) \cdot \log(1 - \interactLabelRawPred).
\end{equation}

$\losslabel$ employs binary cross-entropy to align the final predicted object interaction states $\interactLabelPred$ from the next state decoding module with the ground truth interaction states $\interactLabelGT$:
\begin{equation}
    \losslabel = \interactLabelGT \cdot \log(\interactLabelPred) + (1 - \interactLabelGT) \cdot \log(1 - \interactLabelPred).
\end{equation}

\begin{figure*}[t]
   \centering
   \includegraphics[width=0.9\textwidth]{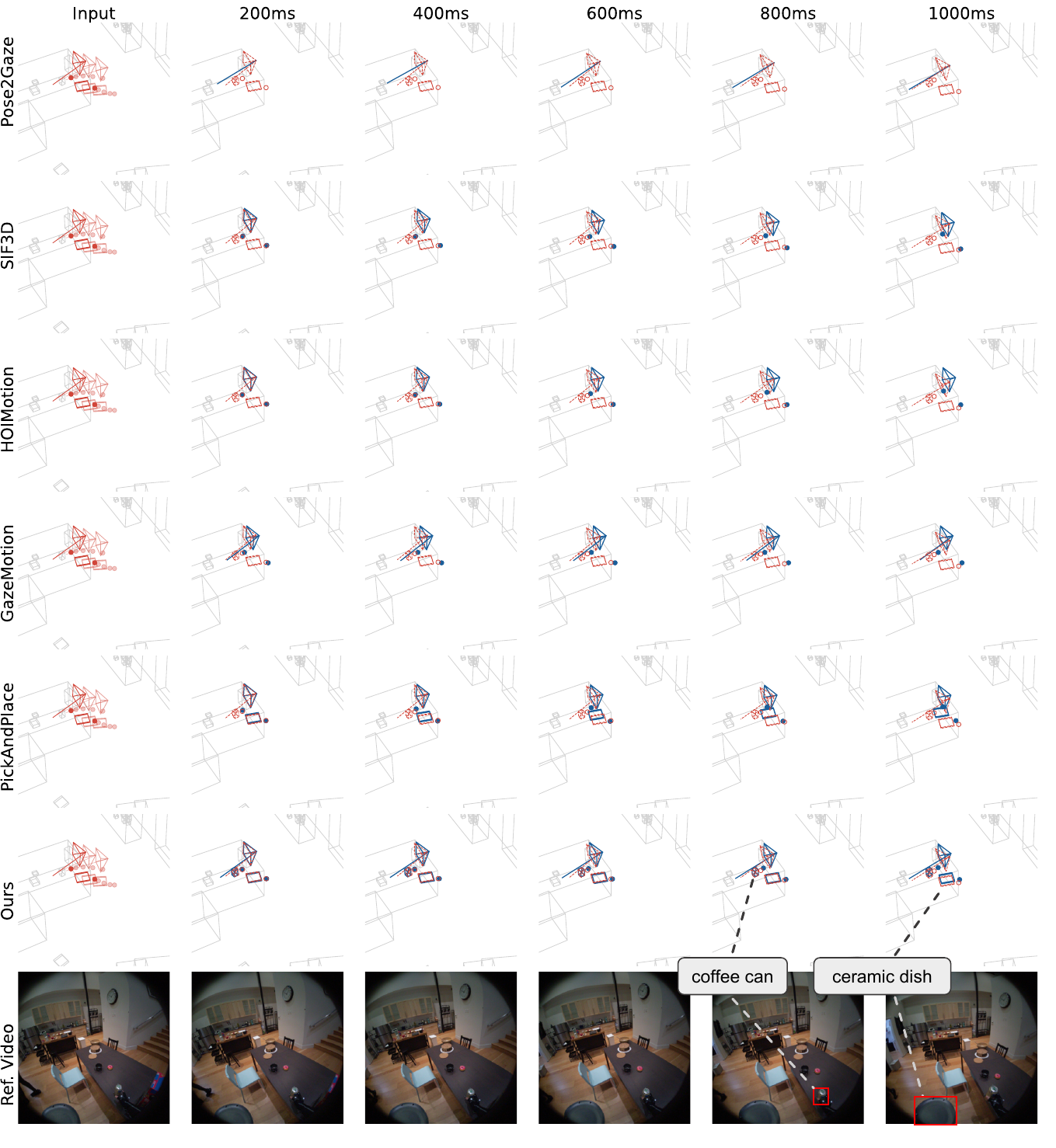} 
   \caption{\textbf{Additional qualitative comparison of our method with state-of-the-art methods on ADT \cite{pan2023aria} dataset.} The leftmost column shows the historical input observation, where color shades from light to dark represent progression from earlier to later time, while the remaining columns display predictions at 200ms intervals from 200ms to 1000ms. In each visualization, red and blue elements represent \ac{gt} and predictions, respectively. The visual elements include: gaze direction (ray), head position and orientation (pyramid), hand positions (two points), and interacted objects (bounding boxes). The interaction object type is labeled in the final frame. The bottom row shows reference frames from the first-person perspective video.} 
   \label{fig:sup_sota}
 \end{figure*}

 \begin{figure*}[t]
   \centering
   \includegraphics[width=0.82\textwidth]{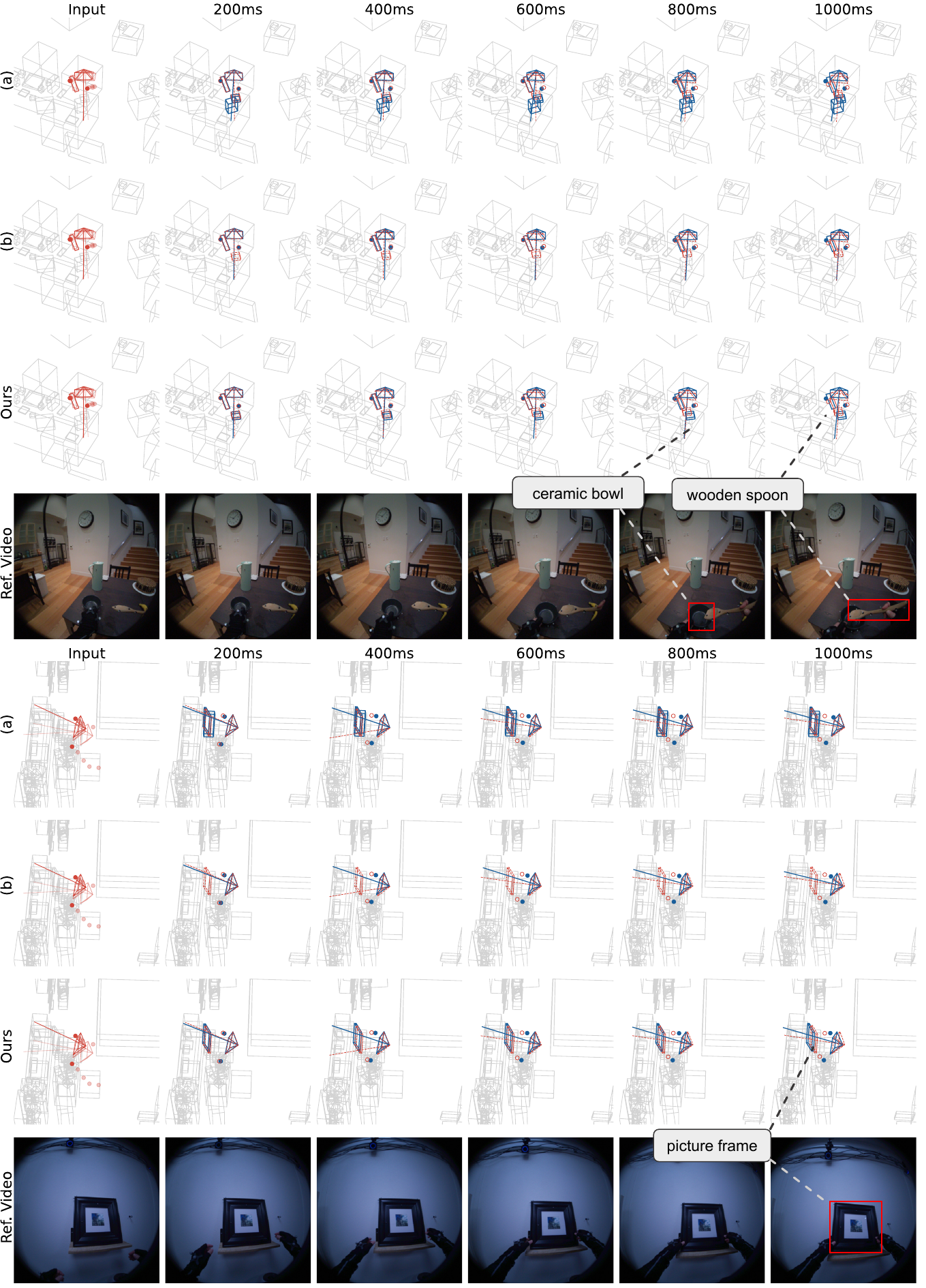} 
   \caption{\textbf{Additional qualitative comparison of our full model with ablated baselines on ADT \cite{pan2023aria} dataset.} Ablation (a) replaces the dynamic GCN with vanilla GCN by removing the dynamic weight matrix $\AdjD$. Ablation (b) removes the interaction intention prediction module and directly decodes the next human states and object interactions. The visualization format is kept the same as in \cref{fig:sup_sota}.} 
   \label{fig:sup_ablate}
   \vspace{-0.3cm}
 \end{figure*}

\section{Experiments}
\label{supp:sec:exp}

\subsection{Baselines}
\label{sup:baseline}

To the best of our knowledge, no existing baselines perfectly align with our specific task. Therefore, we select the most-related works  \cite{hu24gazemotion, hu2024hoimotion, hu2024pose2gaze, lou2024multimodal, razali2022using} and adapted them for a fair comparison.  The detailed implementations are as follows.

\begin{itemize}[leftmargin=1pc]
    \item Pose2Gaze \cite{hu2024pose2gaze}: Pose2Gaze \cite{hu2024pose2gaze} defines the task of pose-based eye gaze prediction. Given a sequence of head directions $H\in \mathbb{R}^{3\times T}$ (3D unit vectors) and body joint positions $P\in \mathbb{R}^{3\times J\times T}$ over $T$ steps, it can predict gaze sequence $G\in \mathbb{R}^{3\times T'}$ for the future $T'$ frames. We replace the full-body pose input with the trajectories of the head and both hands, so $J$ becomes $3$ and $P$ becomes size $3\times 3 \times T$, while all other inputs and outputs remain unchanged. We also match the frame length and frame rate to ours: 15 input frames and 15 output frames at 15 fps. Keeping the model architecture unchanged, we retrain and test Pose2Gaze \cite{hu2024pose2gaze} on the ADT dataset \cite{pan2023aria}.

    \item GazeMotion \cite{hu24gazemotion}: Continuing the line of Pose2Gaze \cite{hu2024pose2gaze}, GazeMotion \cite{hu24gazemotion} defines gaze-guided human motion forecasting task. Given past $T$ frames of 3D human pose $P\in \mathbb{R}^{T\times 3\times J}$ and corresponding eye-gaze unit vectors $G\in \mathbb{R}^{3\times T}$, GazeMotion \cite{hu24gazemotion} first predicts future gaze $G'\in \mathbb{R}^{T'\times 3}$, then utilizes this gaze information to predict future pose $P'\in \mathbb{R}^{T' \times 3\times J}$. Similarly, we adjust both its pose input and output to trajectories of human head and two hands (i.e. $J=3$), and match the frame length and frame rate to ours. The model is retrained and evaluated on the ADT dataset \cite{pan2023aria}.

    \item HOIMotion \cite{hu2024hoimotion}: HOIMotion \cite{hu2024hoimotion} leverages surrounding objects in the environment to aid in future human pose prediction. Specifically, given a historical sequence of body poses $P \in \mathbb{R}^{3\times J\times T}$, head orientations $H \in \mathbb{R}^{3\times T}$ (3D unit vectors), and 3D bounding boxes of scene objects (3D positions of the bounding box's eight vertices $o\in\mathbb{R}^{3\times 8\times M}$), it predicts future pose $P'\in \mathbb{R}^{3\times J\times T'}$ using a GCN-based encoder-residual-decoder architecture. 
    We also adapt its pose representation to trajectories of human head and hands, setting $J$ to $3$, with its other input modalities remaining unchanged.  We conduct retraining and evaluation on ADT dataset \cite{pan2023aria}.

    \item SIF3D \cite{lou2024multimodal}: Unlike HOIMotion \cite{hu2024hoimotion}, SIF3D \cite{lou2024multimodal} utilizes 3D point clouds to represent information about the surrounding environment. It uses this point clouds data $S\in \mathbb{R}^{N\times 3}$, along with historical pose $P\in \mathbb{R}^{3\times J\times T}$ and gaze points sequences $G\in\mathbb{R}^{3\times T}$ within the cloud, to predict future pose $P'\in \mathbb{R}^{3\times J\times T'}$.  Similar to the aforementioned baselines, we simplify the representation of pose by using the trajectories of the person's head and hands, keep other modalities unchanged, retrain and evaluate this model on ADT dataset \cite{pan2023aria}.

    \item PickAndPlace \cite{razali2022using}: The goal of PickAndPlace \cite{razali2022using} is to utilize human pose and gaze to predict an object of interest, and subsequently, the future pose. More specifically, given human pose $P\in\mathbb{R}^{3\times J\times T}$ and gaze data $G\in \mathbb{R}^{3\times T}$ over past $T$ frames, as well as the labels and coordinates of $N$ surrounding objects, PickAndPlace \cite{razali2022using} outputs a score for each object, representing the probability that a human will pick it up within next $T'$ frames. It can also predict the future human pose $P'\in \mathbb{R}^{3\times J\times T'}$. Following the aforementioned baselines, we modified the human pose data in the input and output, replacing it with the trajectories of the person's head and hands. We also redefine the meaning of the object score that this model outputs for each object, so that it represents the probability of the person interacting with the object within the next $T'$ frames, consistent with the setting of our task. After these minor modifications, we retrain and test the model on the ADT dataset \cite{pan2023aria}.
\end{itemize}

\subsection{Implementation Details}
\label{supp:detail}
Our training loss hyperparameters are set to $\lambda_{\text{gaze}}=5.0$, $\lambda_{\text{rot}}=5.0$, $\lambda_{\text{pos}}=1.0$, $\lambda_{\text{center}}=10.0$, $\lambda_{\text{vel}}=1.0$, $\lambda_{\text{int}}=1.0$, and $\lambda_{\text{state}}=1.0$. We train our model using Adam optimizer \cite{kingma2015adam} with an initial learning rate of 0.01 and a step learning rate scheduler with a decay factor of 0.95 per epoch. And the training is performed on 2 NVIDIA Tesla V100 GPUs for 50 epochs with a batch size of 128, taking approximately 4 hours to converge.

\subsection{Additional Qualitative Results}
We present additional qualitative results comparing our method with state-of-the-art baselines (\cref{fig:sup_sota}) and ablated baselines (\cref{fig:sup_ablate}). These results clearly demonstrate the effectiveness of our cognition-based hierarchical decoding framework for situated human behavior prediction tasks. Please refer to the attached anonymous project page in the supplementary material for more video results.

In the input part of the clip represented in \cref{fig:sup_sota}, the subject's left hand is holding a dish, and both the head and left hand trajectories are moving closer to the coffee can on the table. Our model correctly identifies not only that the dish in the left hand will maintain its interaction but also that the right hand will interact with the can. PickAndPlace, however, only determines that the plate in the left hand will continue to interact and fails to predict the interaction with the coffee pot. Furthermore, our model surpasses other baselines in predicting head and hand trajectories.

In the two cases illustrated in \cref{fig:sup_ablate}, ablation study (a), where the dynamic GCN is replaced with a vanilla GCN, resulted in omissions in the prediction of object interaction. Conversely, ablation study (b), where the interaction intention prediction module is removed, led to the prediction of incorrect objects. Our full model, in addition to predicting the correct next active object, also successfully predicts trajectories and gaze closely resemble the ground truth.